\def\year{2021}\relax
\documentclass[letterpaper]{article} 
\usepackage{aaai21}  
\usepackage{times}  
\usepackage{helvet} 
\usepackage{courier}  
\usepackage[hyphens]{url}  
\usepackage{graphicx} 
\usepackage{graphicx}  
\usepackage{natbib}  
\usepackage{caption} 
\usepackage{amsmath}
\usepackage{mathtools,amssymb}
\nocopyright
\usepackage{tikz}
\newcommand*\circled[1]{\tikz[baseline=(char.base)]{
            \node[shape=circle,draw,inner sep=0.6pt] (char) {#1};}}
\usepackage{makecell}
\usepackage{multirow, array}
\usepackage{subfig}
\usepackage[switch]{lineno}

\usepackage{xspace}

\usepackage{color}
\usepackage{pifont}

\title{Real-Time Execution of Large-scale Language Models on Mobile}
\author {
    Wei Niu \thanks{These authors contributed equally} \textsuperscript{\rm 1},
    Zhenglun Kong  \footnotemark[1] \textsuperscript{\rm 2},
    Geng Yuan \textsuperscript{\rm 2},
    Weiwen Jiang \textsuperscript{\rm 4},
    Jiexiong Guan \textsuperscript{\rm 1},
    Caiwen Ding \textsuperscript{\rm 5},
    Pu Zhao \textsuperscript{\rm 2},
    Sijia Liu \textsuperscript{\rm 3},
    Bin Ren \textsuperscript{\rm 1},
    Yanzhi Wang \textsuperscript{\rm 2} \\
}
\affiliations {
    \textsuperscript{\rm 1} College of William and Mary, 
    \textsuperscript{\rm 2} Northeastern University, 
    \textsuperscript{\rm 3} MIT-IBM Watson AI Lab, IBM Research, 
    \textsuperscript{\rm 4} University of Notre Dame, 
    \textsuperscript{\rm 5} University of Connecticut
    
    \textsuperscript{\rm 1} wniu@email.wm.edu, 
    \textsuperscript{\rm 2} kong.zhe@northeastern.edu, \textsuperscript{\rm 2} yuan.geng@northeastern.edu, 
    \textsuperscript{\rm 4} wjiang2@nd.edu, 
    \textsuperscript{\rm 1} jguan@email.wm.edu, 
    \textsuperscript{\rm 5} caiwen.ding@uconn.edu, 
    \textsuperscript{\rm 2} zhao.pu@northeastern.edu, 
    \textsuperscript{\rm 3} sijia.liu@ibm.com, 
    \textsuperscript{\rm 1} bren@cs.wm.edu, 
    \textsuperscript{\rm 2} yanz.wang@northeastern.edu
}

\begin{document}
\maketitle
\begin{abstract}
Pre-trained large-scale language models have increasingly demonstrated high accuracy on many natural language processing (NLP) tasks. However, the limited weight storage and computational speed on hardware platforms have impeded the popularity of pre-trained models, especially in the era of edge computing. 
In this paper, we seek to find the best model structure of BERT for a given computation size to match specific devices. We propose the first compiler-aware neural architecture optimization framework. Our framework can guarantee the identified model to meet both resource and real-time specifications of mobile devices, thus achieving real-time execution of large transformer-based models like BERT variants.
We evaluate our model on several NLP tasks, achieving competitive results on well-known benchmarks with lower latency on mobile devices. Specifically, our model is 5.2$\times$ faster on CPU and 4.1$\times$ faster on GPU with 0.5-2\% accuracy loss compared with BERT$_{\mathrm{BASE}}$. 
Our overall framework achieves up to 7.8$\times$ speedup compared with TensorFlow-Lite with only minor accuracy loss.
\end{abstract}
\section{Introduction}
Pre-trained large-scale language models such as BERT~\cite{devlin2018bert}, XLNet~\cite{yang2019xlnet}, RoBERTa~\cite{liu2019roberta}, and GPT-2~\cite{Radford2019LanguageMA} have substantially advanced the state-of-the-art across a wide spectrum of NLP tasks. 
With the increasing popularity of mobile AI applications and the concerns of information security and privacy, it is desirable to deploy these well trained models on edge devices, and furthermore, to meet real-time requirements. However, these models often consist of hundreds (or even thousands) of layers and hundreds of millions of parameters. 
Therefore, how to accommodate the large and extremely deep models, such as BERT to edge device becomes an imminent problem. 

There have been some efforts to compress BERT model while maintaining the accuracy for downstream NLP tasks. MobileBERT~\cite{Sun_2020} is able to reduce the memory requirement, but there is still a considerable execution overhead due to the large number of model layers, thus leading to high inference latency.
Moreover, the large number of model layers also brings challenges in compiling models to mobile devices. To the best of our knowledge, only TensorFlow-Lite (TFLite)~\cite{TensorFlow-Lite} supports deploying BERT models on mobile CPU (not on mobile GPU), while no other frameworks can even support BERT models on mobile CPU. 

In this paper, we propose the Compiler-Aware Neural Architecture Optimization framework to search for the best BERT structure for mobile devices.
This is the first framework that involves compiler optimizations in the NAS loop, aiming to co-optimize the model accuracy and computation resource usage.
The architectures generated from the framework can be compiled to target mobile devices for real-time execution. The contributions are summarized as follows:

\begin{itemize}
    \item We advance a compiler-aware neural architecture optimization framework to search for a desirable architecture for BERT models to achieve a good balance between accuracy and latency.
    \item We propose a highly effective layer fusion method to reduce intermediate results to achieve better performance on both mobile CPU and GPU.
    \item We evaluate our framework on multiple BERT variants, and compare with a state-of-the-art framework, TFLite, proving that our framework outperforms TFLite by up to 7.8$\times$ speedup. Particularly, This is the first framework supporting BERT execution on both mobile CPU and GPU.
\end{itemize}

Evaluation results show that our models can achieve significantly lower latency with minor accuracy loss. Specifically, our model (executed on our framework) is 5.2$\times$ faster on CPU and 4.1$\times$ faster on GPU with 0.5-2\% accuracy loss compared with BERT$_{\mathrm{BASE}}$. These results demonstrate that our framework can achieve real-time execution of large transformer-based models on an off-the-shelf mobile phone. 


\section{Motivation}

\begin{figure}[t!]
    \centering
    \includegraphics[width=0.95 \columnwidth]{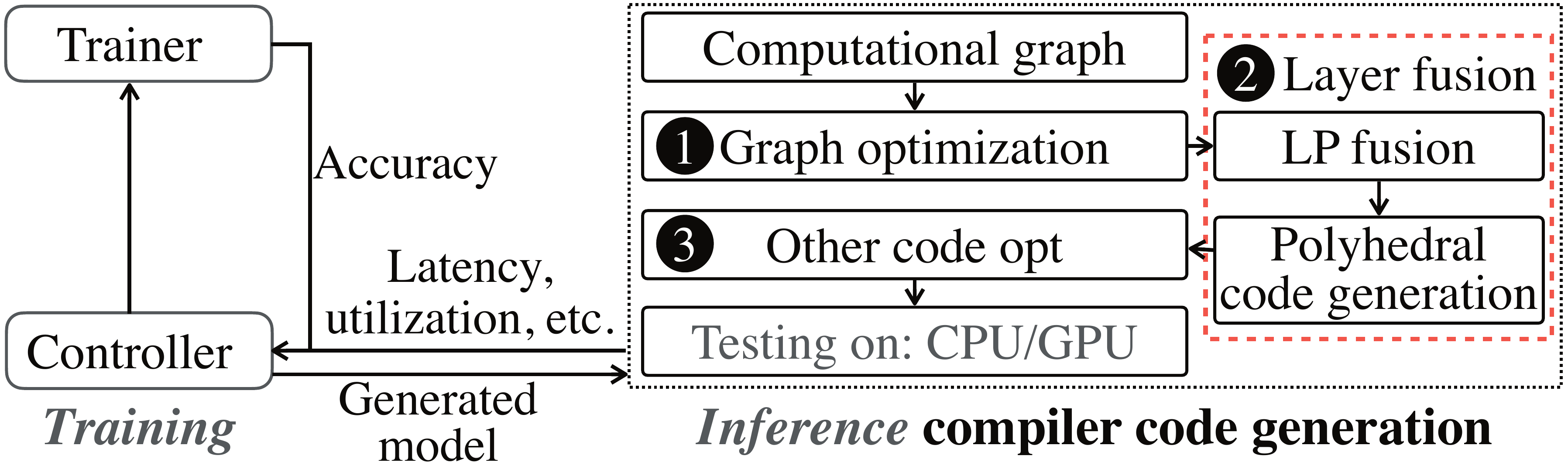}
    \caption{
        Overview of compiler-aware neural architecture optimization.    }\label{fig:nas-compiler-overview}
\end{figure}

Current transformer-based language models have hundreds of millions of parameters. Many of them are extremely deep, leading to high inference latency on edge devices. Take BERT (BERT$_{\mathrm{BASE}}$) as an example. Its high memory and computation cost makes it hard to be deployed on edge devices with limited resource. 
MobileBERT addresses this issue by designing a new model based on BERT$_{\mathrm{LARGE}}$ and distilling it to a small one with the size of 23\% of BERT$_{\mathrm{BASE}}$.
However, MobileBERT's layer count remains the same as BERT$_{\mathrm{LARGE}}$ (over 1,000 computation layers). 
As a result, although MobileBERT has much fewer FLOPs ($19\%$) compared to VGG-16~\cite{simonyan2014deep}, it still runs 6.2$\times$ slower than VGG-16 on mobile, far from real-time execution. Other compact models (e.g., DistilBERT and TinyBERT) remove over half number of layers compared to BERT$_{\mathrm{BASE}}$ by using knowledge distillation. These efforts raise a few open questions:
\begin{itemize}
\item Does BERT really need more layers? Under similar FLOPs, which type of models show higher accuracy on downstream tasks, wider ones or deeper ones?
\item If deeper models are preferred, how can we accelerate the inference to achieve real-time execution?
\end{itemize}



\section{Compiler-aware Neural Architecture Optimization Framework}

\subsection{Overview}
Although the hardware-aware NAS has been proposed to optimize network architectures with the awareness of latency; however, there is still a missing link between neural network search and compiler optimization.
For instance, all the existing hardware-aware NAS: MnasNet~\cite{tan2018mnasnet}, FBNet~\cite{wu2018fbnet}, ProxylessNAS~\cite{cai2018proxylessnas} assumes a general, non-optimized compiler.
It may be fine for computer vision applications with shallow layers, but for the network with hundreds of layers, the inference latency can easily exceed the target without the optimization of the compiler, rendering the hardware-aware NAS useless.
In this work, we involve the compiler optimizations in the NAS search loop, and propose the first compiler-aware neural architecture optimization framework.
Our framework can guarantee the identified model to meet both resource and real-time specifications of mobile devices, thus achieving real-time execution of large transformer-based models like BERT variants while maintaining accuracy.
Our framework consists of two processes: {\it training} and {\it compiler code generation} (as shown in Fig.~\ref{fig:nas-compiler-overview}). 
The training process consists of a controller for predicting/generating the model hyperparameters (i.e., network architecture), and a trainer to train the predicted model and (quickly) evaluate its accuracy by fine-tuning the model to downstream tasks.
The compiler code generation process takes the predicted model and returns execution information (e.g. latency, number of fused layers, CPU/GPU utilization).
The execution information together with the model accuracy from the training process will be feedback to the controller to improve the prediction of neural architectures. After the compiler-aware NAS, the generated codes by our optimized compiler will be deployed for mobile CPU/GPU executions.

\begingroup
\setlength{\tabcolsep}{1.0pt} 
\begin{table}[t!]
\centering
\small{
\begin{tabular}{|c|c|c|c|c|c|}
    \hline
     Model & MRPC & STS-B & RTE & CoLA & \makecell{Latency \\ CPU/GPU}\\
    \hline
    BERT$_{\mathrm{BASE}}$ & 88.9 & 85.8 & 66.4 & 52.1 & 257/186 \\
    \hline
    DistilBERT & 85.0 & - & 65.5 & 51.3 & 145/133 \\
    \hline
    MobileBERT & 88.8 & 84.4 & 66.2 & 50.5 & 73/69\\
    \hline
    BERT(ours) w/o distill. & 84.9 & 81.6 & 63.8 & 45.7 & 60/54 \\
    \hline
    BERT(ours)  & 88.5 & 83.8 & 65.8 & 49.7 & 60/54 \\
    \hline
    BERT(ours) with NAS & 88.4 & 83.5 & 65.6 & 49.2  & \textbf{49/45} \\
    \hline
\end{tabular}
}
\caption{Evaluation results on GLUE benchmark. MRPC, STS-B, RTE, and CoLA columns show accuracy, and the last column shows inference latency on mobile CPU and GPU (with a unit of ms). All models are optimized with layer fusion and code generation (i.e., they already run faster than their TFLite implementation) with a fixed sequence length of 128. MobileBERT and BERT(ours) are trained with knowledge distillation, while BERT(ours) w/o distill. is trained directly from a deep-narrow structure.}
\label{tab:eva_distil}
\end{table}
\endgroup

\section{Experiments}

\begingroup
\setlength{\tabcolsep}{5.5pt} 
\begin{table*}[t!]
\centering
\small{
\begin{tabular}{|cc|c|cccc|cccc|}
     \hline
     Framework & \multirow{2}{*}{\#FLOPs} & TFLite & \multicolumn{4}{c|}{Our framework (without layer fusion)} & \multicolumn{4}{c|}{Our framework (with layer fusion)} \\
     Device   & ~ & \makecell{CPU} & \makecell{CPU} & Speedup & \makecell{GPU} & Speedup & \makecell{CPU} & Speedup & \makecell{GPU}  & Speedup \\\hline
     DistilBERT with NAS   & 10.9G  & 188ms & 157ms & 1.2$\times$ & 237ms & 0.8$\times$ & \textbf{105ms} & \textbf{1.8$\times$} & \textbf{86ms}  & \textbf{2.2$\times$}   \\ \hline
     BERT$_{\mathrm{BASE}}$ with NAS & 21.8G & 352ms & 276ms & 1.3$\times$ & 412ms & 0.9$\times$ & \textbf{196ms} & \textbf{1.8$\times$} & \textbf{147ms} & \textbf{2.4$\times$} \\ \hline
     BERT(ours) with NAS             & 4.6G  & 98ms  & 89ms  & 1.1$\times$ & 152ms & 0.6$\times$ & \textbf{49ms}  & \textbf{2.0$\times$} & \textbf{45ms}  & \textbf{2.2$\times$} \\ \hline
\end{tabular}
}
\caption{Inference latency comparison of our framework and TFLite on mobile CPU and GPU, demonstrating effectiveness of layer fusion. All models are generated with English Wikipedia dataset. TFLite does not support BERT on mobile GPU.}
\label{tab:BERT-performance-mobile}
\end{table*}
\endgroup

\subsection{Methodology}

\noindent{\bf Models and datasets.} 
We test our framework on three mainstream BERT models: BERT$_{\mathrm{BASE}}$~\cite{devlin2018bert}, DistilBERT~\cite{sanh2019distilbert}, and MobileBERT~\cite{Sun_2020}.
For pre-training, we use the same corpus as the original BERT model: BooksCorpus~\cite{zhu2015aligning} and English
Wikipedia datasets~\cite{devlin2018bert}. We fine-tune the pre-trained models on GLUE benchmark~\cite{wang2018glue}.

\noindent{\bf Evaluation setup.}
Our training is executed on GPU-AI (Bridges GPU Artificial Intelligence) nodes on the Extreme Science and Engineering Discovery Environment (XSEDE)~\cite{xsede}. 
We use two node types: Volta 32 and Volta 16.
We conduct the experiments using HuggingFace Transformer toolkit~\cite{wolf2019huggingface}.

We evaluate our framework on a Samsung Galaxy S20 cell phone with Qualcomm Snapdragon 865 which consists of a Qualcomm Kryo 585 Octa-core CPU and a Qualcomm Adreno 650 GPU. We use a Samsung Galaxy S10 with a Qualcomm Snapdragon 855 that consists of a Kryo 485 Octa-core CPU and an Adreno 640 GPU, and an Honor Magic 2 with a Kirin 980 that consists of an ARM Octa-core CPU and a Mali-G76 GPU for portability evaluation. For each model, we run our framework and TFLite 100 times with 8 threads on CPU and all pipelines on GPU. 


\subsection{Accuracy and Latency Results}


We compare the accuracy and latency of six models: BERT$_{\mathrm{BASE}}$, MobileBERT, DistilBERT, BERT(ours) w/o distillation, BERT(ours), and BERT(ours) with NAS. We apply layer fusion to all BERT variants to show the effectiveness of compiler-aware model optimization.

BERT(ours) w/o distillation is directly trained with a 28-transformer block deep-and-narrow structure; BERT(ours) is derived from further distilling from a teacher model. 
Note that BERT(ours) uses the same distillation method as MobileBERT.
For BERT(ours) with NAS, 200 training epochs are used for the overall NAS.
The models are evaluated on four downstream tasks: MRPC, STS-B, RTE, and CoLA.  Accuracy and latency results are shown in Table ~\ref{tab:eva_distil}, with the optimizations of our proposed layer fusion. 
We can see that BERT(ours) improves accuracy by 3-4\% compared to BERT(ours) w/o distillation under the same latency. 
All of our three models can achieve notably lower latency compared to BERT$_{\mathrm{BASE}}$, DistilBERT, and MobileBERT.

By further applying our compiler-aware NAS, which is our BERT(ours) with NAS model, we manage to significantly reduce latency compared to BERT$_{\mathrm{BASE}}$, DistilBERT, and MobileBERT on both CPU and GPU. Compared with BERT$_{\mathrm{BASE}}$, our model is 5.2$\times$ faster on CPU and 4.1$\times$ faster on GPU with 0.5-2\% accuracy loss.
Compared with MobileBERT, our model is 1.49 $\times$ faster on CPU and 1.53$\times$ faster on GPU with only 0.4-1\% accuracy decrease.


\subsection{Effectiveness of Compiler Optimizations}

Table~\ref{tab:BERT-performance-mobile} shows inference latency comparison results. The fully optimized framework can achieve up to 2.0$\times$ speedup on CPU, and 2.4$\times$ on GPU, over TFLite's CPU execution. Notably, comparing to BERT$_{\mathrm{BASE}}$ on TFLite, our framework (BERT(ours) with NAS on GPU) can achieve up to $7.8\times$ speedup.
Without compiler optimizations, our baseline implementation runs slightly better than TFLite on CPU, because TFLite is already optimized for BERT models. 
Without compiler optimizations, GPU performance is unusually worse than CPU (only 0.6$\times$ speedup for BERT(ours) over TFLite on CPU). This is because extremely deep layers generate many intermediate data, while mobile GPU memory performs worse than CPU due to its smaller and simpler cache hierarchy. 



\section{Conclusion}

This paper presents a compiler-aware neural architecture optimization framework to search for the best BERT structure for mobile devices. The proposed framework guarantees the identified model to meet both resource and real-time specifications of mobile devices, achieving real-time execution of large transformer-based models (like BERT and its variants). The proposed framework achieves up to $7.8\times$ speedup over TFLite.   
Our BERT model generated from the framework outperforms both BERT$_{\mathrm{BASE}}$ and MobileBERT with small accuracy loss on popular NLP downstream tasks.

\newpage
\bibliography{template} 

\end{document}